\documentclass{article}

\usepackage[nonatbib,final]{NFFL2021}

\usepackage[utf8]{inputenc} 
\usepackage[T1]{fontenc}    
\usepackage{hyperref}       
\usepackage{url}            
\usepackage{booktabs}       
\usepackage{amsfonts}       
\usepackage{amsmath,amssymb}
\usepackage{amsthm}
\usepackage{adjustbox}
\usepackage{wrapfig}
\usepackage{subcaption}
\theoremstyle{plain}

\theoremstyle{definition}

\usepackage{nicefrac}       
\usepackage{microtype}      
\usepackage{xcolor}         
\usepackage{enumitem}
\usepackage{todonotes}
\usepackage{booktabs}
\usepackage{multirow}
\usepackage[normalem]{ulem}
\useunder{\uline}{\ul}{}
\usepackage[square,numbers]{natbib}
\title{Contribution Evaluation in Federated Learning: Examining Current Approaches}

\author{%
  Vasilis Siomos \\
  Imperial College London\\
  City University London\\
  \texttt{vasilis.siomos@city.ac.uk}
   \And
   Jonathan Passerat-Palmbach \\
   Imperial College London \\
   Flashbots \\
   \texttt{j.passerat-palmbach@imperial.ac.uk}
}

\begin{document}

\maketitle

\begin{abstract}
Federated Learning (FL) has seen increasing interest in cases where entities want to collaboratively train models while maintaining privacy and governance over their data. In FL, clients with private and potentially heterogeneous data and compute resources come together to train a common model without raw data ever leaving their locale. Instead, the participants contribute by sharing local model updates, which, naturally, differ in quality. Quantitatively evaluating the worth of these contributions is termed the Contribution Evaluation (CE) problem. We review current CE approaches from the underlying mathematical framework to efficiently calculate a fair value for each client. Furthermore, we benchmark some of the most promising state-of-the-art approaches, along with a new one we introduce, on MNIST and CIFAR-10, to showcase their differences. Designing a fair and efficient CE method, while a small part of the overall FL system design, is tantamount to the mainstream adoption of FL.
\end{abstract}

\section{Introduction}
Federated Learning \cite{mcmahan2017communicationefficient} has emerged as an appealing paradigm of collaborative model training in the digital privacy age. Its appeal stems from allowing distributed computation on fragmented, privately owned data without sacrificing privacy, or performance by utilising the exchange of gradients/model weights instead of raw data. Initially developed to target massive numbers of low-resource edge devices \cite{gboard}, it has since found application in real-world cross-institutional settings \cite{melloddy_2021,healthchain_project}, too. FL creates a path for the formation of consortia between competing companies since, under privacy guarantees, if the global model is better than the individual ones, participating companies stand to benefit. These two settings are commonly referred to as cross-device and cross-silo, respectively \cite{kairouz_advances_2019}.

\textit{Fairness} in FL comprises two notions: the first, analogous to regular ML fairness, is ensuring the model is not biased against any group of participants. Here, we focus on the second, often disregarded, notion of fairly compensating clients for their contributions. In mainstream FL research, it is ubiquitous to assume all clients unconditionally want to participate and remain in the federation, meaning their compensation is de-coupled from the training process (e.g., participants in Google’s Gboard deployment received no external reward apart from the model \cite{gboard}, while the MELLODDY \cite{melloddy_2021} consortium participants receive arbitrary -- from the perspective of the model -- EU funding). Realistically, however, \textit{potential} clients have the agency to decide whether to join and remain in the federation based on how they, and their collaborators, will be rewarded for it. In cross-device FL, reluctance to participate is justified by the personal nature of the data, meaning users are reluctant to share it altruistically, even under privacy guarantees. In cross-silo FL, participants take a business risk and incur data curation and training costs and thus expect fair compensation. This fact, coupled with the smaller number of participants, all of which are available and persist throughout training, mean CE is both more feasible and more needed in cross-silo settings, and thus this is where most current work focuses. 

The problem of assigning a numeric value to what participants share with the federation is termed Contribution Evaluation (CE), and is part of the problem of designing the overall mechanism that motivates client participation, called Incentive Mechanism Design. A \textit{payoff scheme} which guarantees clients are compensated in fair proportion to their contributions is necessary to convince them to join the federation. CE is closely related to similar evaluation problems in data valuation and feature attribution for explainable AI, treating data as a commodity, and creating data marketplaces, so understanding current FL approaches necessarily involves examining non-federated data valuation methods. However, CE in FL introduces additional challenges, which we will aim to highlight.
\section{Game Theory background}
The mathematical framework used in data/contribution evaluation is borrowed from Coalitional Game Theory (CGT), well-suited to model client interactions and how they lead to the formation of coalitions of cooperating parties. CE can thus be framed as solving a game $G$ with $N$ participating clients, where outcomes consist of two elements. The first is a partitioning of the players into disjoint coalitions, each of which is mapped to a real number representing its value (also called \textit{utility}) using a \textit{characteristic function} $v$. Most often, $v$ is chosen to be the coalitional model's accuracy on a held-out validation set. The second is a payoff vector, $\mathbf{x}=(x_1,...,x_n)$, distributing the utility of each coalition to its participants. Generally, $x_i$ are non-negative, and the L1 norm of $\mathbf{x}$ is only upper-bounded by the utility $\left(\sum_{i \in C^j} x_i \leq v(C^j) \forall j \in {1,...k}\right)$; if the equality holds it is called \textit{efficient}. If an efficient payoff vector satisfies the \textit{individual rationality} condition of $x_i \ge v(\{i\}) \forall i \in N $, it is called an \textit{imputation}. Intuitively, an imputation is a set of payoffs that exactly splits the value of the coalition to each member in a manner in which it is sensible for all of them to remain in the coalition.

A solution concept $\sigma$ is a mapping from $G$ to a set of such outcomes, given a notion of our desired goal, such as fairness, or ensuring a single coalition, including all clients (called the grand coalition), forms. Such outcomes of $G$ (if they exist), are called rational under $\sigma$. Fair CE consists then of solving $G$ for fairness, choosing a fairness-oriented $\sigma$, and computing its rational outcomes, resulting in the set of payoff vectors that fairly represent the worth of participants' contributions to the federation. How to best incorporate the numeric values $\mathbf{x_i}$ in the FL system design is the purview of Incentive Mechanism Design and outside our scope.

$G$ is \textit{superadditive} if $v(C \cup D) \ge v(C) + v(D) \ \forall \ \text{disjoint} \ C,D\subseteq N $, leading the grand coalition to always form, since it is always non-detrimental to grow the coalition. In fact, unconditionally assuming the grand coalition forms is valid \textit{only} under this property, otherwise solving for \textit{stability}, i.e., convincing clients to join and stay in our federation, instead of fairness, must be considered. The following well-known solution concepts each focus on one of these two goals.

The \textit{Core} is the set of all imputations under which no coalition would receive a higher payoff by leaving the grand coalition ($x(C) \ge v(C) \  \forall  \ C \subseteq N$). It ensures the stability of the grand coalition and includes all \textit{realistic} payoffs, as imputations outside it are not viable if players are free to leave the grand coalition. However, the Core is not guaranteed to be non-empty or unique, for the imputations that lie in it (if they exist), no preference is defined, and they might be unfair for some players. Note that, even then, a player's Core payoff is still higher than their unitary utility. 

If instead, we desire to calculate a fair measure for the value of $G$ from the perspective of each client, we can use their marginal contribution to a coalition $S$ to calculate their worth; calculated either post-addition, as $v(S)-v(S \backslash \{i\})$, or pre-addition, as $\left(v\left(S\cup\{i\}\right)-v\left(S\right)\right)$. Furthermore, we can consider contributions of individual data points: the well-known statistical Leave-One-Out (LOO) metric is the post-addition marginal contribution of a point. In our setting, the LOO is intractable, but marginal contributions are sometimes referred to as client-level LOO. Alone, the marginal contribution does not adjust for the order in which participants join the coalition, or for the size of the coalition at that point. In contrast, Shapley \cite{Shapley_value} defined four axioms of fairness that he believed a fair measure, i.e. a payoff vector, $\phi$ should satisfy:
\begin{enumerate}[itemsep=0pt,leftmargin=*]
    \item \textbf{Efficiency/Pareto Optimality}: $\sum_{i \in N} \phi_i(v)=v(N)$
    \item \textbf{Symmetry}: If $v(S \cup \{i\})=v(S \cup \{j\}) \ \forall S\subseteq N \backslash \{i,j \}$, then $\phi_i(v)=\phi_j(v)$
    \item \textbf{Null Player}: If $v(S \cup \{i\})=v(S) \ \forall S\subseteq N \backslash \{i \}$, then $\phi_i(v)=0$
    \item \textbf{Additivity}: If $u$ and $v$ are characteristic functions, then $\phi(u + v) = \phi(u) + \phi(v)$.
\end{enumerate}

The only \cite{ichiishi1983game} payoff vector satisfying them all is called the Shapley Value (SV) \cite{Shapley_value}. For a coalition $N$ and the set of all characteristic functions $\mathcal{V}$ on $2^N$, the Shapley Value $\phi$ on $\mathcal{V}$ is given by \cite{peleg2007}:
$$
\phi_i(v) = 
\sum_{S\subseteq N\backslash\{i\}}
\frac{|S|!(n-|S|-1)!}{n!}
\left(v\left(S\cup\{i\}\right)-v\left(S\right)\right) \quad \forall v \in \mathcal{V}, i \in N
$$

We must emphasize the conceptual difference between the Core and the SV. To calculate the latter, we assume all the players have agreed to join the game, or, equivalently, that the game is super-additive, and we take on the role of each of the players, calculating the fair value of $G$ according to their marginal contributions, producing $\phi_i$. Instead the Core includes the viable payoffs a central entity can deliver to each player to keep them in the grand coalition.

\section{Existing contribution evaluation approaches}
Designing  a fair CE method for FL can be broken down to the following problems:

\textbf{Choosing a solution concept and efficiently computing the imputation(s) of $G$ under it}. Directly calculating either the SV or the Core is exponentially complex $O(2^N)$ w.r.t to the number of clients. Even in the less crowded cross-silo setting, each calculation requires fully training the underlying model using the data from the sampled client subset, meaning finding exact solutions is practically intractable for models with long training times, like neural networks.

\textbf{Adapting to FL}. The entire theoretical framework based on CGT, and the resulting methods, were first developed to tackle data valuation. Contrary to that setting, in FL, model updates are being shared instead of raw data, which are potentially obfuscated/encrypted for privacy, and are only becoming available to the server sequentially at every round. Thus, adapting methods is non-trivial and in some instances impossible, as we will see.

\textbf{Balancing utility and valuation accuracy}. CE is, at its core, a data valuation problem, yet FL minimizes/eliminates the exchange of data. The more a CE method burdens clients with calculations or the more data it requires from them, the less compatible it becomes with the overall FL pipeline, while on the other hand FL obfuscation measures like secure aggregation \cite{bonawitz2017practical} inhibit accurate valuation.

\subsection{Evaluation using the Shapley Value}
If we accept Shapley's axioms as a comprehensive set of desirable properties for a CE metric, then the SV is unique in satisfying them and thus the de facto metric to calculate. Since its exact computation is intractable, several data valuation methods focus on efficiently and accurately approximating it. These are not all directly transferable to FL, in fact, most are not, but nevertheless understanding them is necessary to move to FL adaptations. The seminal work in \cite{ghorbani_data_2019} presented Data Shapley, an approximation based on Monte Carlo sampling; by substituting the \(1/n\) constant with \(1/n!\) in the definition of SV, and using the uniform distribution over all possible permutations of the data $\Pi$, approximating the SV equates to finding an expected value, a problem lending itself to sampling-based solutions:
$\phi_i'  = \frac{1}{n!}\sum_{S \subseteq N/\{i\}} \frac{V(S \cup \{i\}) - V(S)}{ { n-1 \choose |S|}}= \mathop{\mathbb{E}}_{\pi \sim\Pi}\left[ V(S_{\pi}^{i}\cup\{i\})-V(S_{\pi}^{i})\right]$, where $S_\pi^i$ is the set of data points coming before datum i in permutation $\pi$. Sampling permutations $\pi \sim \Pi$ and sequentially evaluating the marginal contributions of each sample i to $S_\pi^i$, before adding them to the list of training data, leads, after many iterations, to average marginal contributions which are unbiased estimates of the SVs. The intuition behind Data Shapley is that, when using the performance on a validation/test set $\hat{T}$ as the characteristic function, this set will always be finite and drawn from the true test distribution $T$, and thus, a bootstrapped approximation of the SV up to the intrinsic observation noise that comes from drawing $\hat{T}\sim T$ is satisfactory. The Distributional SV \cite{ghorbani_distributional_2020} stabilises the SV w.r.t. perturbations of the training set; whereas the SV is only valid for a single draw of training samples, the Distributional SV is an unbiased estimate of the value of adding a datum to any dataset drawn from the underlying distribution. A similar repertoire of approximations was proposed in \cite{jia_towards_2019}, most interesting of which is using the combinatorial technique of group-based testing, with group sampling chosen to allow efficient approximations of the pairwise differences of SVs between clients, from which the individual SV approximations can then be efficiently recovered. 

These algorithms need all the data to be simultaneously available, a prerequisite not satisfied in the cross-device setting, and one which introduces a big computational and communication overhead, in the cross-silo one. Furthermore, in both settings, Shapley's symmetry axiom becomes a hindrance for any federation where not every client joins simultaneously: in that case `early adopter` participants should be rewarded more, for helping bootstrap the global model. Using the SV, not only is that not accomplished, but a client can even join the process late, with a local dataset that is a duplicate of another, and his SV payoff would be the same as the early adopter with the same dataset.

Recognising this shortcoming, and extending/modifying the SV for FL, an approach termed Federated Shapley was proposed in \cite{wang_principled_2020}: Let $I=\{1,...,N \}$ denote the set of participants selected by the coordinator during a T-round FL process. Let $I_t$ be the set of participants selected in round t, $I_{j:k}$ the selection sets between rounds $j,k$ and $I_t \subseteq I$. Then, the federated SV of participant $i$ at round $t$ is defined as:
$$
s_t^v(i)=\frac{1}{|I_t|} \sum_{S\subseteq I_t \backslash \{i\}}\frac{1}{ {|I_t-1| \choose |S|}}\left[ v(I_{1:t-1}+(S\cup \{i\}))-v(I_{1:t-1}+S) \right] \ \ if \ \ i \in I_t, else \ 0,
$$
and the federated SV is their sum across rounds $s^v(i)=\sum_{t=1}^{T}s_t(i)$. So, at every round, for every participant, the average marginal contribution across all possible coalitions is used to get a \textit{temporal snapshot} of the values of participants' datasets. Straight-forward adaptations of the original fairness axioms hold for the federated SV \cite{wang_principled_2020}. Exact computation is even more costly than that of the SV ($O(T2^m)$ complexity, where $m$ the maximum number of participants in a round), and hence MC sampling or group-based testing approximations \cite{jia_towards_2019} are still used in the original publication.

Another FL-compatible SV approximation was presented in \cite{efficientandfair}, which \textit{only} uses the gradients produced during training the grand coalition's model to approximately reconstruct all other models necessary for the SV calculation. Whereas for the exact SV computation we need to retrain a model on all possible combinations of contributing clients, meaning clients expend time and resources on training $2^n-2$ useless models, using this pseudo-model approximation, only the useful model, generated using all participant data, is trained. Pseudo-models are constructed at every round by combining the marginal contributions (pre-addition) of the clients to the global model \textit{of the previous round}. Three distinct algorithms to achieve that are proposed:
\begin{itemize}[itemsep=0pt, leftmargin=*]
    \item One-Round (OR): The server uses the marginal contributions, normalized by local dataset sizes, to update the $2^{N}$ pseudo-models stored at the server at every round. The SV for each client is calculated after training concludes, using the final pseudo-models.
    \item $\lambda$-Multi-Rounds ($\lambda$-MR): A so-called round-CI (originally presented by the same authors in \cite{profit}) is calculated at every round, following the OR procedure, using the latest pseudo-models. After training, round-CIs are weighted by a decay factor which emphasizes earlier contributions, normalized by their per-round sum, and aggregated, to produce the SVs.  
    \item Truncated Multi-Rounds: $\lambda$-MR can be sped up by adding a threshold for the decay $\lambda$, below which updating the pseudo-models stops. The aggregation formula for the SV now also weighs round-CIs proportionally to the test accuracy of the global model at the corresponding round, emphasizing rounds where test accuracy was high.
\end{itemize}
A key connection we make here is noticing that round-CI and Federated Shapley $s_t^v(i)$ refer to the same quantity, albeit the approximations used to calculate each of them differ conceptually. $\lambda$-MR focuses on earlier contributions, which, arguably, are more important than later ones in steering the loss of the global model towards a better minimum, via the time decay coefficient.

\subsection{Alternative Methods of Contribution Evaluation}
While the SV is arguably over-represented in the literature, other solution concepts have not been studied extensively. Recently, Yan and Procaccia \cite{lovecore} examined using the Core and a related solution concept called the Nucleolus for data evaluation. As, like SV, the Core is prohibitively expensive to compute exactly, we can choose to relax it in several ways, allowing for efficient computation. Extending it to include imputations where no actor benefits more from leaving the grand coalition, if they have to pay a cost $\epsilon$ for leaving, we arrive at the $\epsilon$-Core: $C_{\epsilon}(v)=\left\{x\in R^N:\sum_{i\in N}x_i=v(N); \ \sum_{i\in S}x_i \ge v(S)-\epsilon, \forall S \subseteq N \right\}
$. This is a logical relaxation in cross-silo/enterprise FL, where each client will have signed a participation contract before joining, and the expenses of breaking it to leave the federation play the role of $\epsilon$. The \textit{least} core is the solution of the linear program minimizing $\epsilon$:
$$
\text{min} \ \epsilon \ \ \textbf{s.t.} \ \ \sum\nolimits_{i \in N}x_i = v(N) \ \text{  and  } \
    \sum\nolimits_{i \in S}x_i + \epsilon \geq v(S) \quad \forall S \subseteq N
$$
The quantities $v(S) - \sum\nolimits_{i \in S}x_i$ are the deficits of each coalition, expressing its dissatisfaction with the payoff scheme. The least core is obviously non-empty if we let $\epsilon$ be unbounded, while the core itself is non-empty if $\epsilon^* \leq 0$. 
Similarly, the \textit{nucleolus} is the unique imputation which lies in the least core and minimizes the largest deficit. Solving for the least core requires training $2^N-2$ models to calculate the RHS of each constraint, and then solving a linear program with $2^N-1$ constraints and $N+1$ decision variables. In \cite{lovecore} they prove that, if we allow the Core guarantee to be violated by an amount $e$, $\delta$ percent of the time, i.e.,
$
\underset{S \sim D}{Pr}\left[\sum_{i \in S}x_i+\epsilon^*+e \ge v(S) \right] \ge 1-\delta$, then an approximate least core solution, with probability $1-\Delta$, requires sampling only $O\left(\frac{\tau^2(\log{n}+\log(1/\Delta)}{e^2 \delta^2}\right)$
coalitions, where $\tau$ is the difference between the maximum and minimum value of the characteristic function. Unfortunately, they also prove that the same relaxations for the Nucleolus lead to a sampling procedure still exponential in the number of participants.

Other approaches utilize smart contracts on a blockchain to tackle the CE problem in the decentralized FL setting. Blockchain is especially well-suited to the task due to allowing verifiable/transparent transactions between peers, aiding trust. In \cite{cai2020}, two decentralized protocols, for the cross-device and the cross-silo setting respectively, are introduced, to calculate the percentages of final ownership of the model each client should be allotted, based on the sum of the marginal contributions of the data points in each partner's dataset. However, they compute the contribution scores off-chain, in an unverifiable manner. All the centralized CE methods we examined so far implicitly assume the central party is trusted, and thus no type of secure aggregation is necessary. But in the decentralized FL scenario, that is no longer the case, and trust and security become even more important. Secure aggregation \cite{bonawitz2017practical}, a masking technique in FL where client updates are encrypted, so that their sum (which is the only quantity the server normally needs to update the global model) stays the same, but individual updates cannot be examined. With secure aggregation in place, we cannot calculate the SV or any such individual metric. As a compromise, in \cite{ma_transparent_2021}, the authors propose grouping participants and using secure aggregation only inside the groups, before computing, on-chain, the SV for the whole group and subsequently splitting that equally among the group members. An altogether use of Blockchain is proposed in \cite{liu2020fedcoin}, where, while the FL process itself is orchestrated by a central server, the server publishes, as tasks on a consensus blockchain, the calculations needed for CE (chosen to be the exact SV). Miners in the blockchain get paid by either calculating these to create blocks, or by verifying blocks, similar to how the bitcoin network works. While this side-steps approximations, miners are paid from the profits of the federation, meaning less revenue is left for the participants.
\begin{figure}
     \centering
     \begin{subfigure}[b]{0.3\textwidth}
         \centering
         \includegraphics[width=\textwidth]{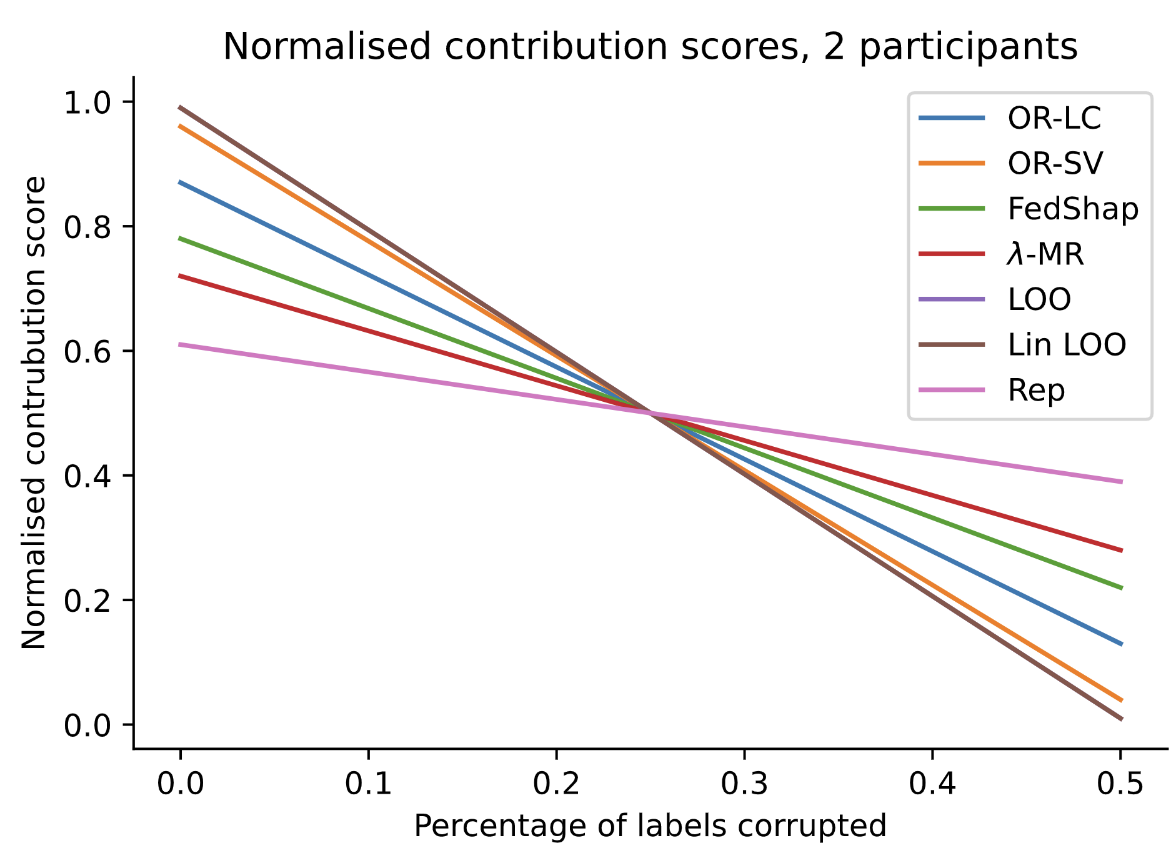}
     \end{subfigure}
     \begin{subfigure}[b]{0.3\textwidth}
         \centering
         \includegraphics[width=\textwidth]{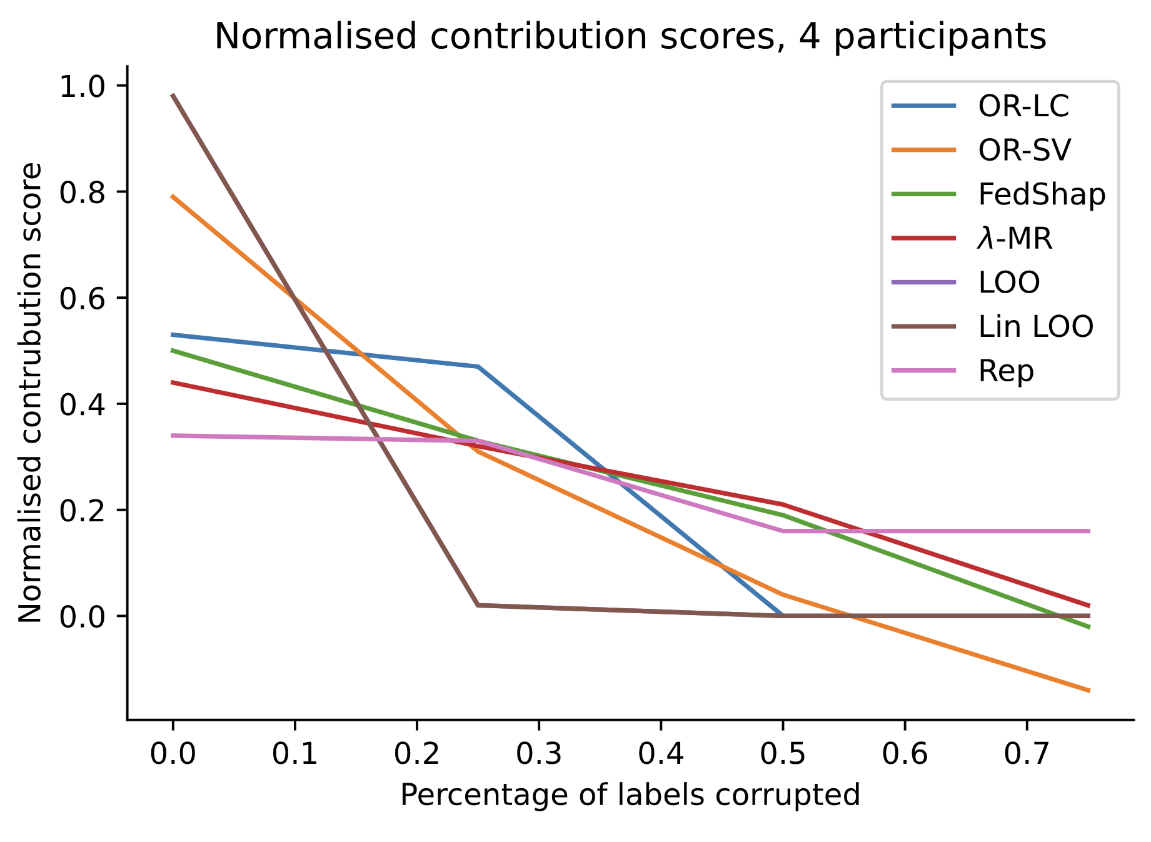}
     \end{subfigure}
     \begin{subfigure}[b]{0.3\textwidth}
         \centering
         \includegraphics[width=\textwidth]{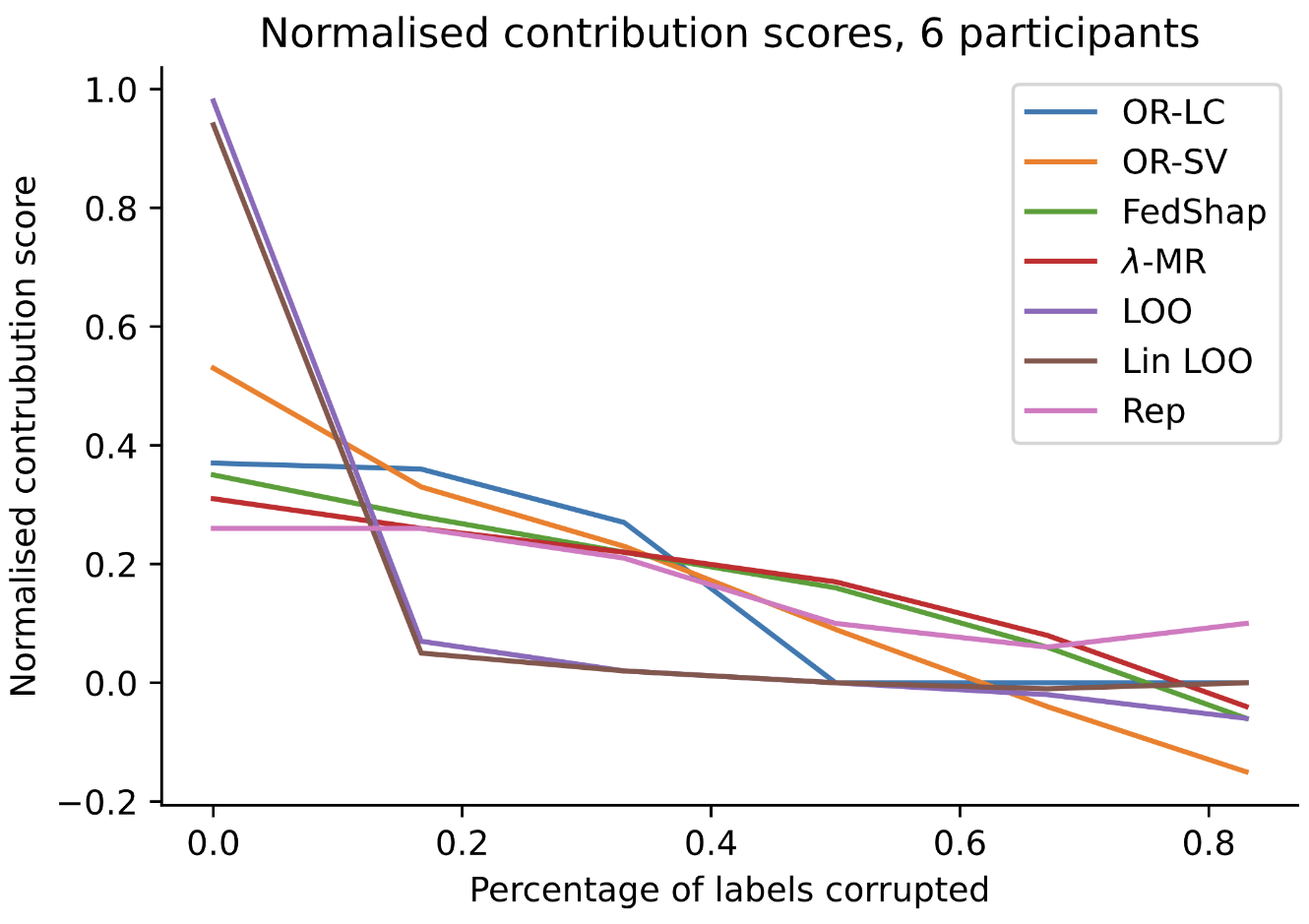}
    \end{subfigure}
          \begin{subfigure}[b]{0.3\textwidth}
         \centering
         \includegraphics[width=\textwidth]{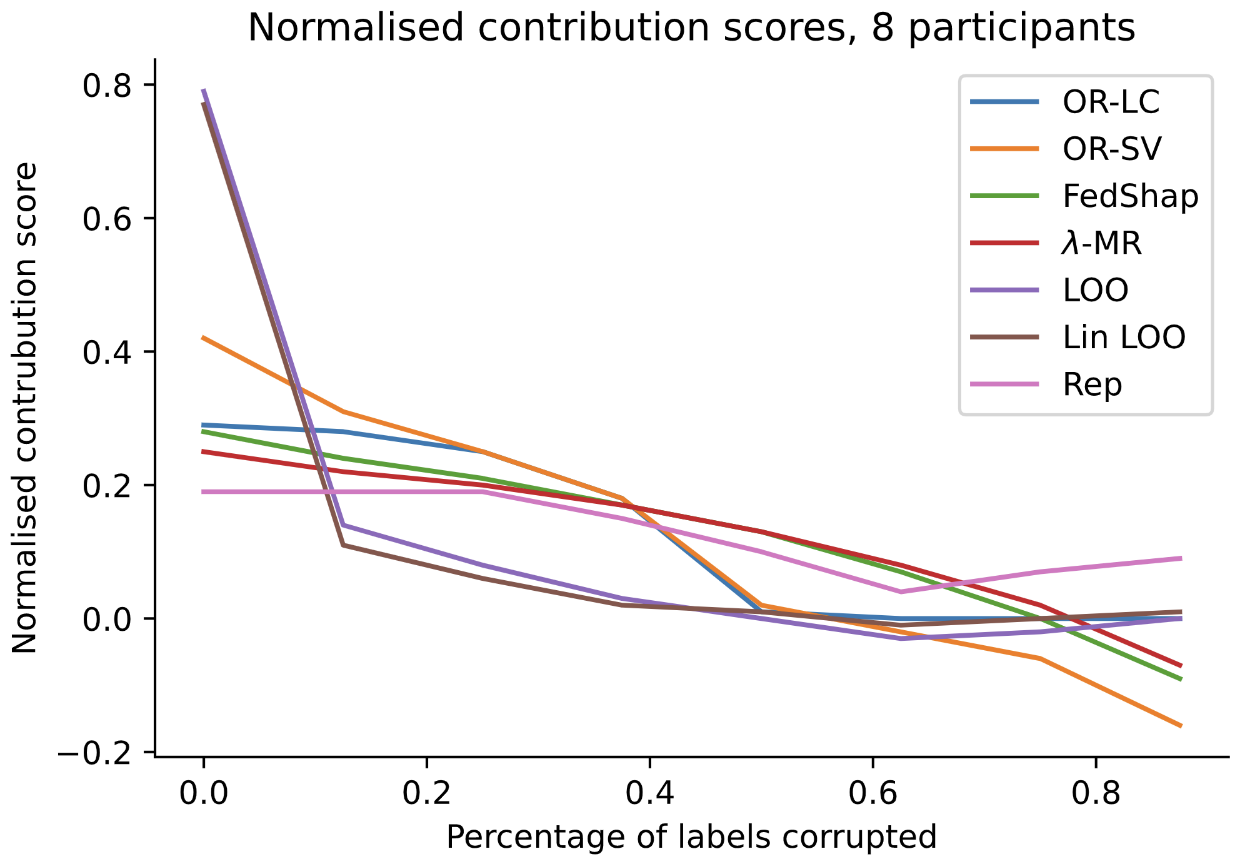}
     \end{subfigure}
    \begin{subfigure}[b]{0.3\textwidth}
         \centering
         \includegraphics[width=\textwidth]{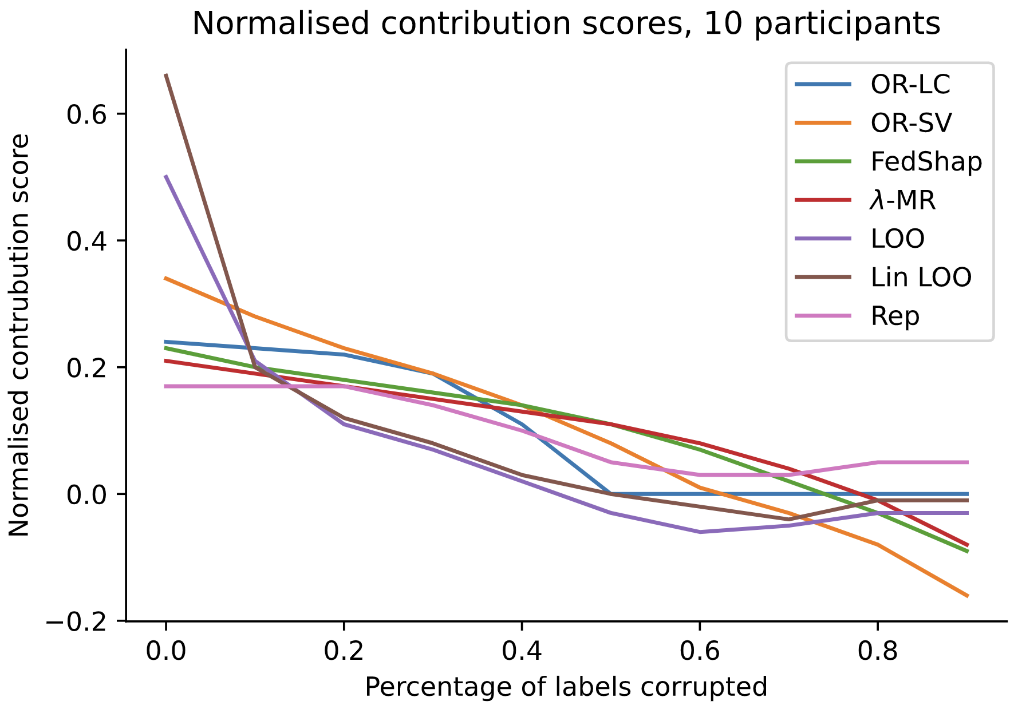}
     \end{subfigure}
     
        \caption{Comparison of normalized payoffs on MNIST}
        \label{fig:mnist}
\end{figure}

\subsection{Models as rewards}
Another crucial aspect of CE is combating \textit{free-riders}: in the context of CGT, a free-rider is a client trying to extract a non-zero payoff, while contributing null, or redundant, information to the coalition. Null updates are tackled by any solution concept satisfying the Null Player fairness axiom, such as the SV, but more sophisticated attacks are not. Free-riding attacks are an understudied problem in FL \cite{lin2019free,fraboni2021free}, but rewarding attackers with a null payoff is not enough when the global model, encapsulating the entire value of the federation, is transmitted to all clients at every round. Especially in cross-enterprise FL, the global model is arguably much more valuable than any single monetary payoff; a free-rider being able to access the global model is an intolerable flaw. 

While free-riding remains an open problem in the global-model paradigm, some recent work moderates the quality of the model that is transmitted to each client, leading to a model-as-reward paradigm; the more a client contributes, the better the model they receive is. Having such a deterrent to free-riders can complement free-rider detection techniques such as STD-DAGMM \cite{lin2019free}, and, as a side-benefit, this solves other conceptual problems in designing an FL Incentive Mechanism, such as enticing non-profit organizations (e.g., public hospitals, university labs) to take part. 

Fairness in this paradigm can be achieved by ensuring the quality of each client's model is in fair proportion to their contribution quality. Such a mechanism, only applicable to Bayesian models, was designed in \cite{sim_collaborative_2020}, where the distributed models have been injected with label noise proportional to the clients' contributions to induce different model quality. In \cite{Lyu2020, lyu_how_2020}, instead of a global model, they propose a gradient-based economy, awarding clients with the gradients of other participants based on a credibility score. Every client converges to a different local model, with the clients uploading the most useful data getting the most gradients and thus ending with the most performant model. In \cite{Lyu2020}, the server keeps copies of the local models, periodically evaluates them on $\mathcal{V}$, and assigns a corresponding credibility score to each client. This so-called reputation of each client is used to assign a fraction of the total gradients to them, with the most reputable one being allowed to download all the gradients. If a client's credibility falls under a threshold, he is ostracized from the set of reputable participants and does not get any updates, thus combating free-riders. In \cite{lyu_how_2020}, they extend this to the decentralized setting, where no public validation set exists, requiring a much more involved design: clients initially publish a differentially private set of local samples, generated by a differentially private GAN, and every other client uses their local model to predict labels for these samples and sends them back to the samples' owner. After comparing these against the owner's predictions, a \textit{local} credibility score is assigned to each client. These are aggregated, and used to decide the set of reputable participants, who train by sending DP gradients in return for tokens. Tokens can be spent to download the gradients of other clients. A participant with good data remains in the reputable set for longer, earning more tokens and thus having access to more gradients. All these transactions are recorded on a blockchain to make the peer-to-peer system verifiable and robust. Arguing against the use of validation performance-based evaluations, \cite{zhang_hierarchically_2020} instead proposed to form client tiers based on publicly verifiable factors, such as the quantity of their data and the cost they incurred to collect it, and giving each tier access to a model trained on a corresponding amount of data.

\section{Benchmarking on MNIST and CIFAR-10}
Here, we aim to --partially-- fill an empirical gap in the related literature by comparing state-of-the-art CE methods for FL for a varying number of clients. The detailed experimental setup and additional experiments are relegated to the appendix. We choose OR-SV, $\lambda$-MR \cite{efficientandfair}, FedShapley \cite{wang_principled_2020}, round-level LOO, both weighted and unweighted \cite{substra} and Reputation \cite{kang_incentive_2019} as the most promising CE methods applicable to FL. Besides these, we noticed that the pseudo-model approximation from \cite{efficientandfair} for OR-Shapley can also be used to efficiently find the Least Core. We name this CE method, OR-LC.

We examine their scalability, and their ability to differentiate contribution quality, by varying the number of clients from 2 to 10, and injecting their datasets with an increasing amount of label noise. Note that other artificial differentiations are possible, such as different dataset sizes/ percentages of data class ownership, but we chose the most intuitive/straightforward one. To provide a non-formal evaluation, we compare them by plotting the normalized contribution scores against the noise rate of each participant: intuitively a good CE method must be able to differentiate different data qualities and exhibit monotonically decreasing behaviour, and the slope's steepness reflects how aggressively each method reduces payoffs for sub-optimal contributions.

Firstly, we experiment on MNIST \cite{mnist}. As seen in fig. \ref{fig:mnist}, all the methods can differentiate the clients' data quality, but they greatly differ as to how austere they are. LOO and linearly weighted LOO heavily favour the best contributor and almost ignore the rest. The payoff vector dictated by Reputation is closer to an egalitarian uniform split, which is expected as it is much coarser than the other methods. The profiles of $\lambda$-MR and FedShapley stand between those, smoothly transitioning from high to low values. OR-SV and OR-LC value highly good and mediocre contributors, but they sharply drop when contribution quality deteriorates past a threshold. 

\begin{figure}[h]
     \centering
     \begin{subfigure}[b]{0.33\textwidth}
         \centering
         \includegraphics[width=\textwidth]{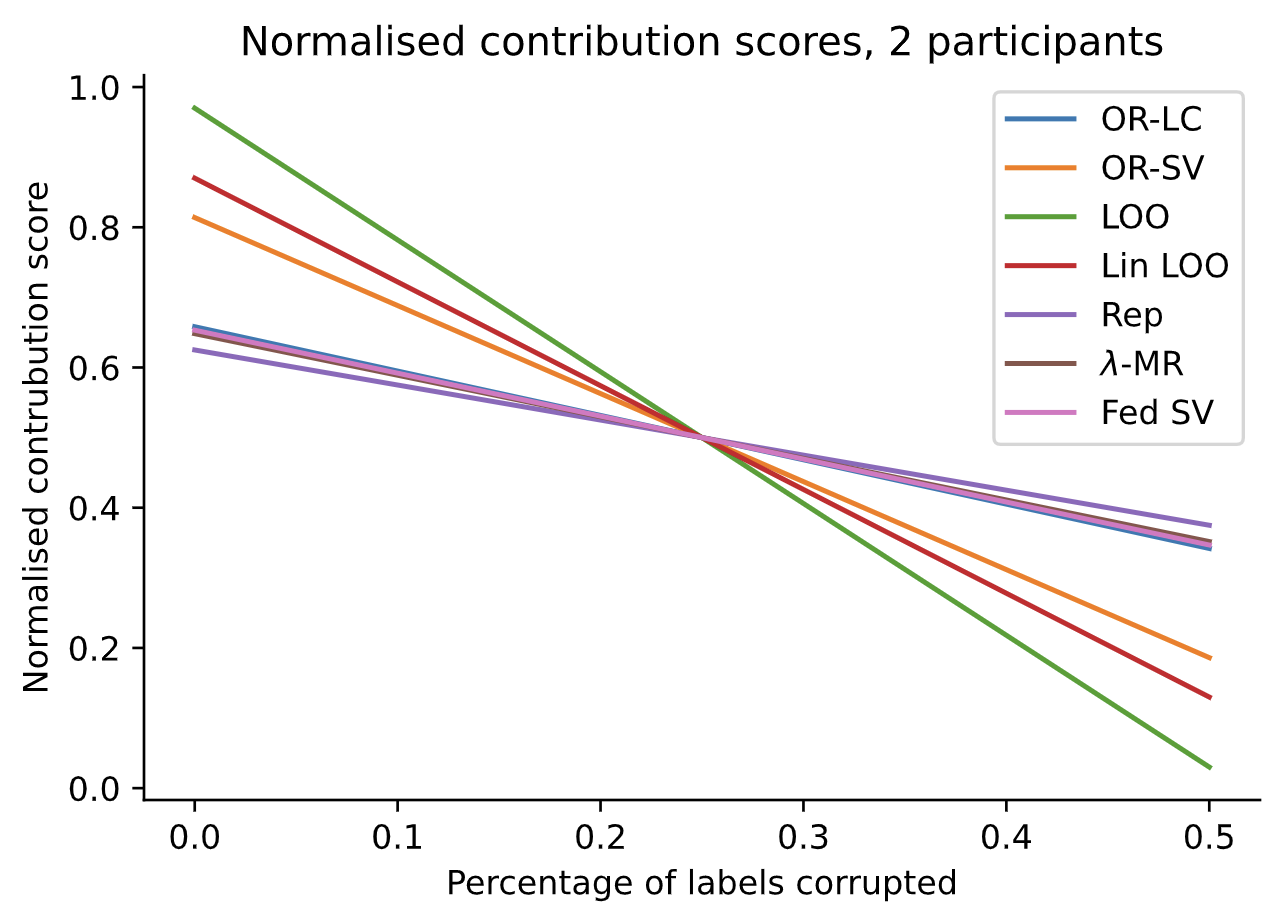}
     \end{subfigure}
     \begin{subfigure}[b]{0.33\textwidth}
         \centering
         \includegraphics[width=\textwidth]{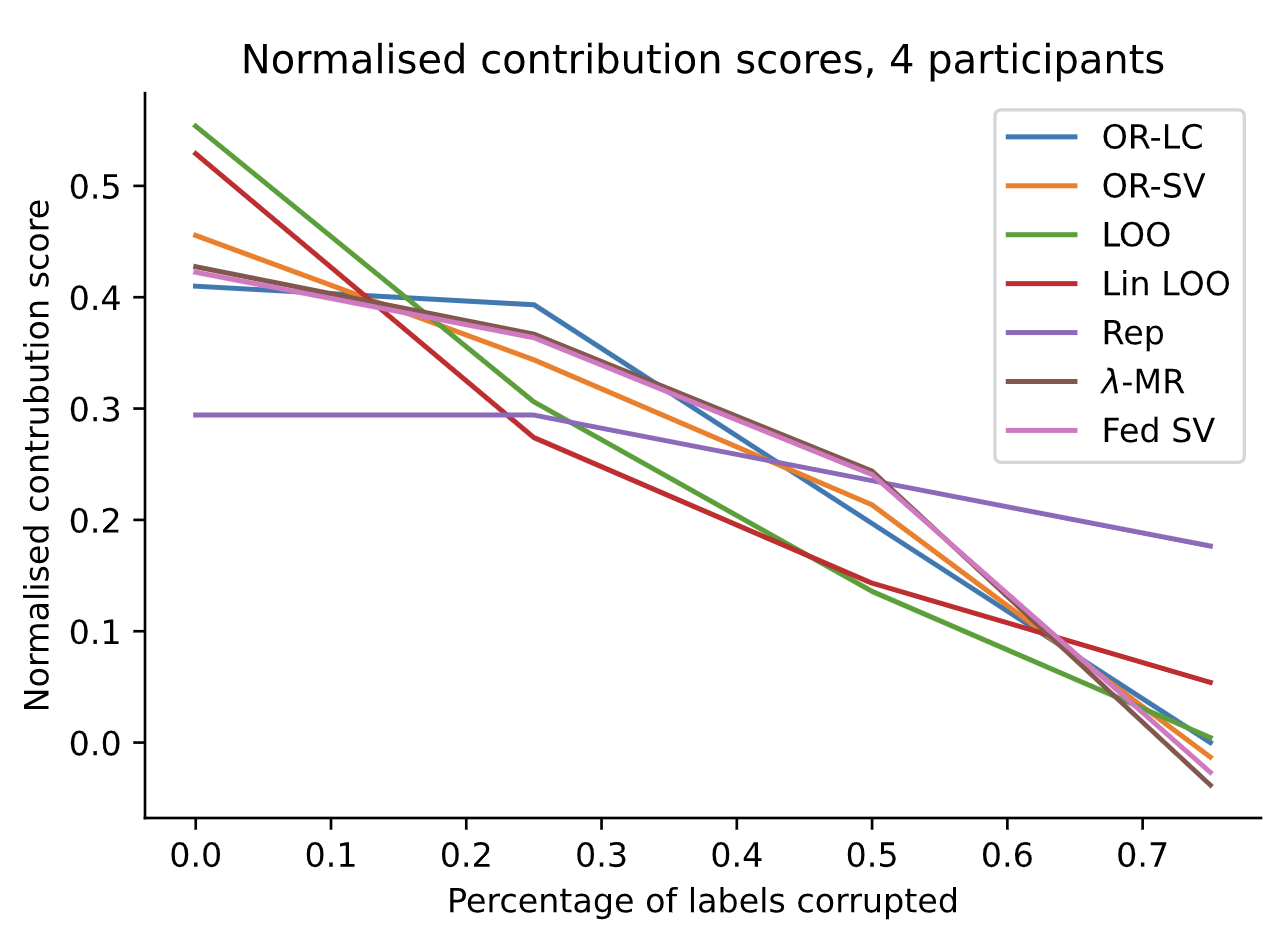}
     \end{subfigure}
     \begin{subfigure}[b]{0.33\textwidth}
         \centering
         \includegraphics[width=\textwidth]{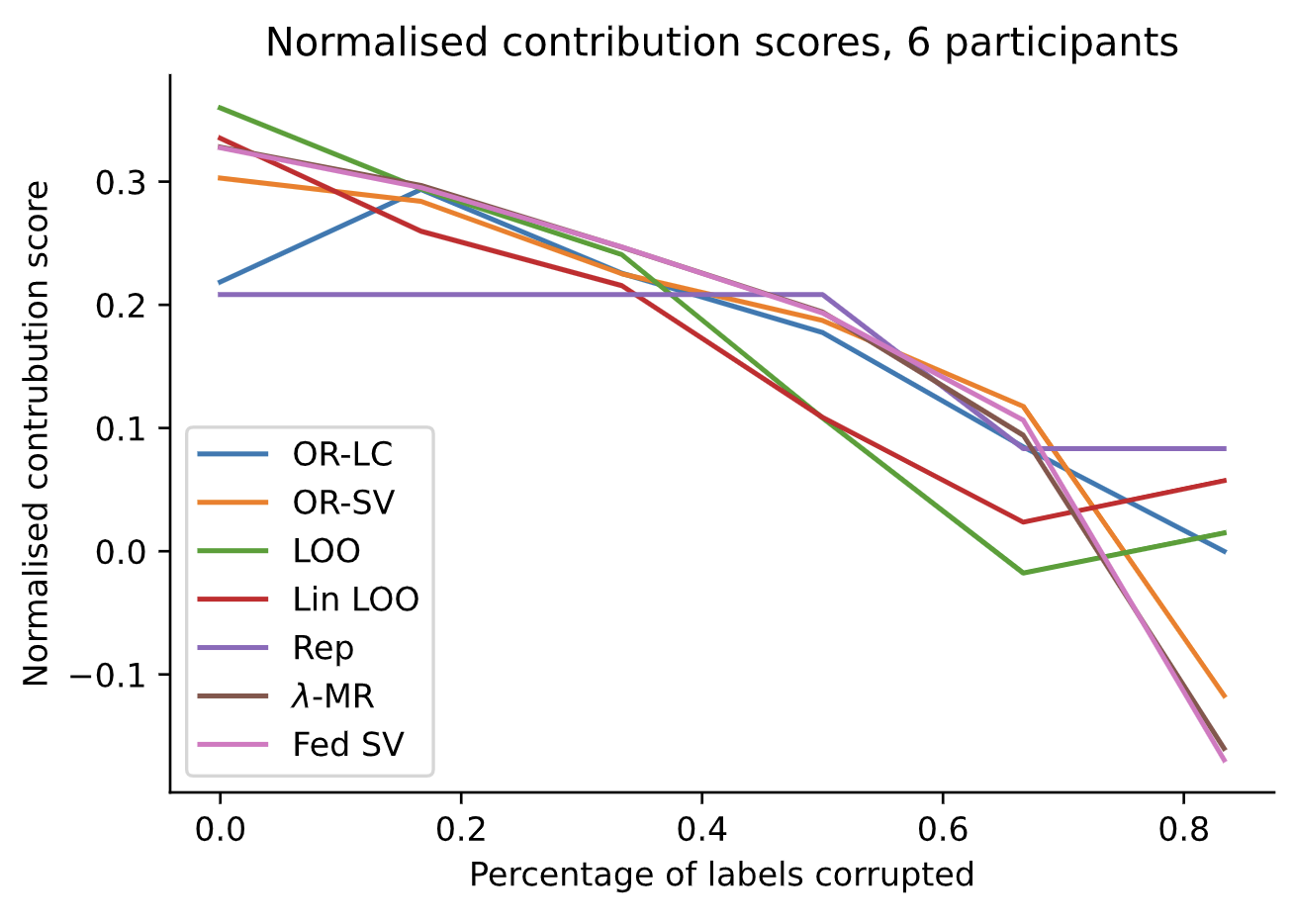}
    \end{subfigure}
          \begin{subfigure}[b]{0.33\textwidth}
         \centering
         \includegraphics[width=\textwidth]{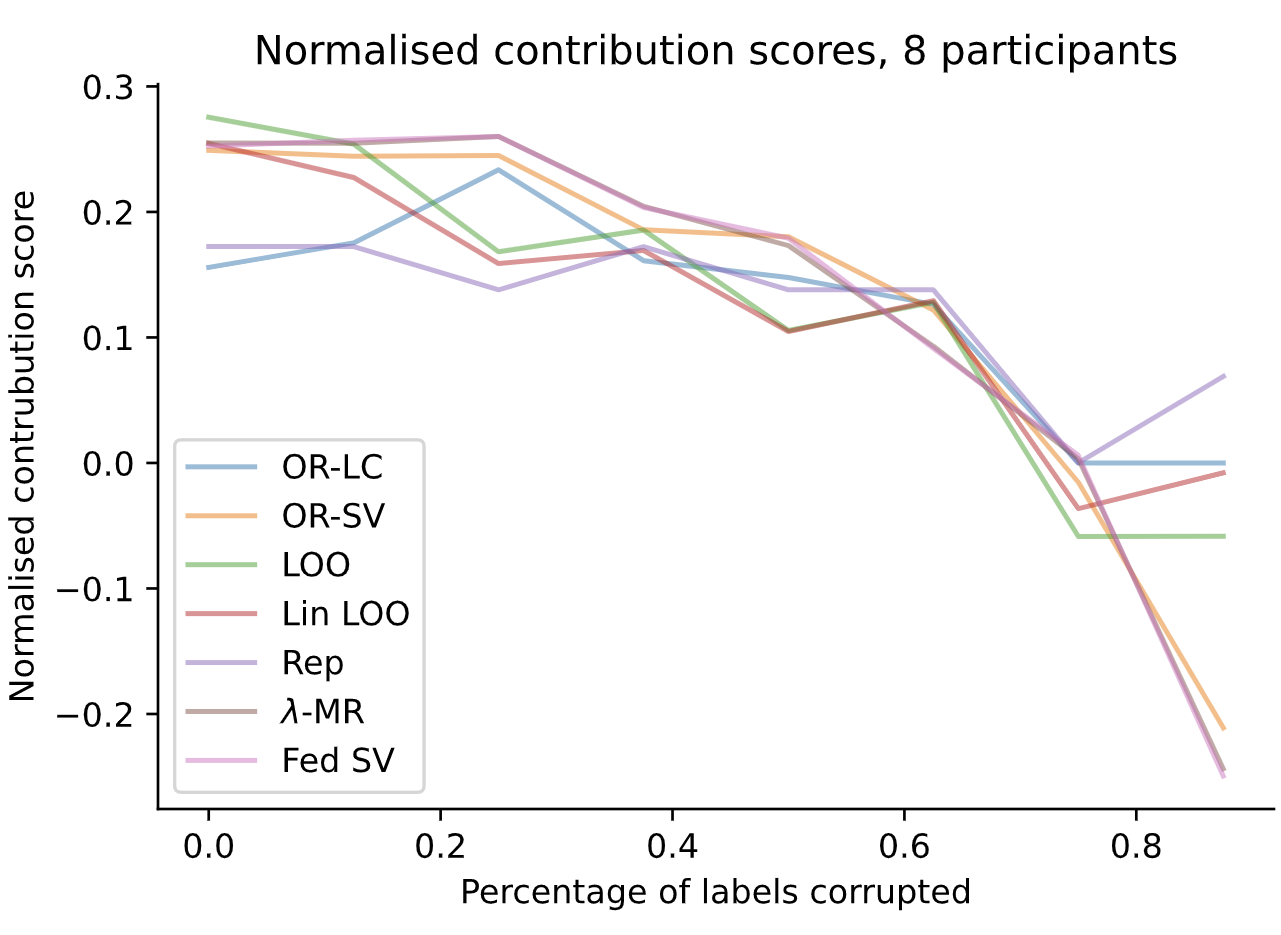}
     \end{subfigure}
    \begin{subfigure}[b]{0.33\textwidth}
         \centering
         \includegraphics[width=\textwidth]{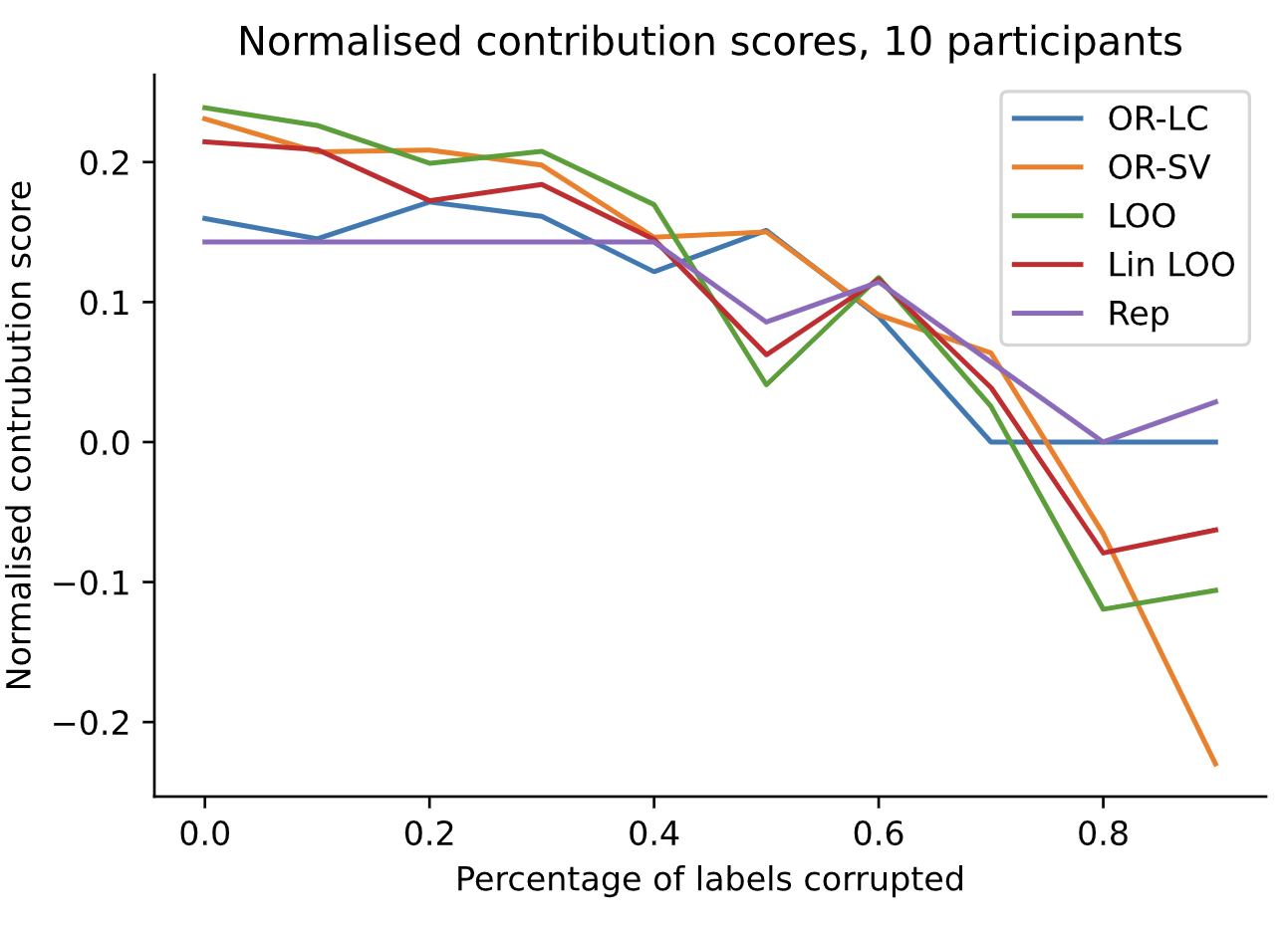}
     \end{subfigure}
     
        \caption{Comparison of normalized payoffs on CIFAR-10. MR for 10 participants requires the server to store 1024 ResNets, causing a crash.}
        \label{fig:cifar}
\end{figure}

Finally, we perform the same experiments on the much harder CIFAR-10\cite{cifar} dataset. Under this more challenging setting, the methods continue to differentiate between data quality, but differences between them become harder to discern as can be seen in fig. \ref{fig:cifar}. Depending on the application, the best suited CE method will differ. To give an example, following the law of supply and demand, the more interested participants there are, the more aggressively the federation can punish sub-optimal contributions.

\section{Conclusion}
In this work, we examined the existing set of approaches to evaluate participants' contributions in FL training in a fair way. Most existing work has focused on adapting the renowned concept of the Shapley Value to work with the idiosyncrasies of Federated Learning. Even at the conceptual level, that presents difficulties, so alternate approaches such as moderating the distributed model quality on a client level are being explored. Drawing from the Computational Game Theory area, we highlighted the related notions of Core and Nucleolus as currently understudied but promising alternatives to the Shapley Value. We then illustrated our careful literature study by comparing how the most promising methods deal with clients with different levels of corrupted labels in both the MNIST and CIFAR-10 classification tasks.

Future work in the area has to tackle the problem of rewarding early contributions more, in contrast to how the Shapley Value works, while also being interpretable as fair by the participants.

\clearpage
\bibliography{references}
\newpage
\appendix

\section{Appendix}

\subsection{Experimental setup}
The injected noise is symmetric, meaning corrupted labels are flipped to any other label uniformly at random, and the noise ratios are linearly spaced between 0 and 1; when using 2 clients, their sets have $[0\%,50\%]$ noise, when using 4 participants the noise rates are $[0\%, 0.25\%, 0.5\%, 0.75\%]$ and so on. Experiments are run with 5 seeds (which remain consistent across the evaluation of different methods), and we only report mean values since variance was negligible. As we are more interested in the cross-silo setting, we assume full client participation in every round.

To compute OR-SV, we use the adjusted version of OR presented in \cite{toefinder}, which leads to a faster and more accurate approximation by rearranging terms in the calculation.

For $\lambda-MR$, $\lambda$ is set to 0.8, per the original paper. Federated Shapley is computed via the same MR approximation, but without the time-decay. To compute round-level LOO, we measure each client's marginal contribution at every round, and sum up the per-round LOO both without weights and multiplied by the value of the current round, i.e., the second-round LOO counts twice as much as the first). The Reputation metric is the average of the Heaviside function applied to the LOO. To arrive at OR-LC, notice that if we approximate all the pseudo-models, we can evaluate them to formulate the LP problem constraints, which is trivial to solve. Since the LC is not unique, we report the first imputation lying in the LC found by the LP solver.

Our setup on MNIST largely follows the setup of \cite{toefinder}. The model used is a two-layer MLP with $64$ hidden units and Dropout ($p=0.5$). No preprocessing is done apart from scaling the images to the $[0,1]$ range. The number of training rounds is set to $5$, and the number of local epochs to $10$. Local optimization uses SGD with momentum $m=0.5$ and $lr=0.01$. 

For the CIFAR-10 experiment, the model used is adapted of the fast ResNet-9 presented in \cite{resnet}; the architecture is the same but without Batch Norm, and optimization is simplified by training the local models using Adam\cite{adam} with learning rate $1e-3$. The rounds and local epochs remain the same as before. The multi-round approximations of $\lambda$-MR and FedShapley need the server to store 1024 ResNet models, causing a crash, even for our lightweight model (but highlighting their memory footprint).
\subsection{Time complexity of experiments}

We can also examine the computational cost for each family of methods across the two datasets and the number of participants. As expected, MR methods are much more expensive and almost exponential to the number of participants due to high inference cost, but for the more complex CIFAR-10 training, the training itself dominates.
\begin{figure}[b]
\centering
    \includegraphics[width=0.5\textwidth]{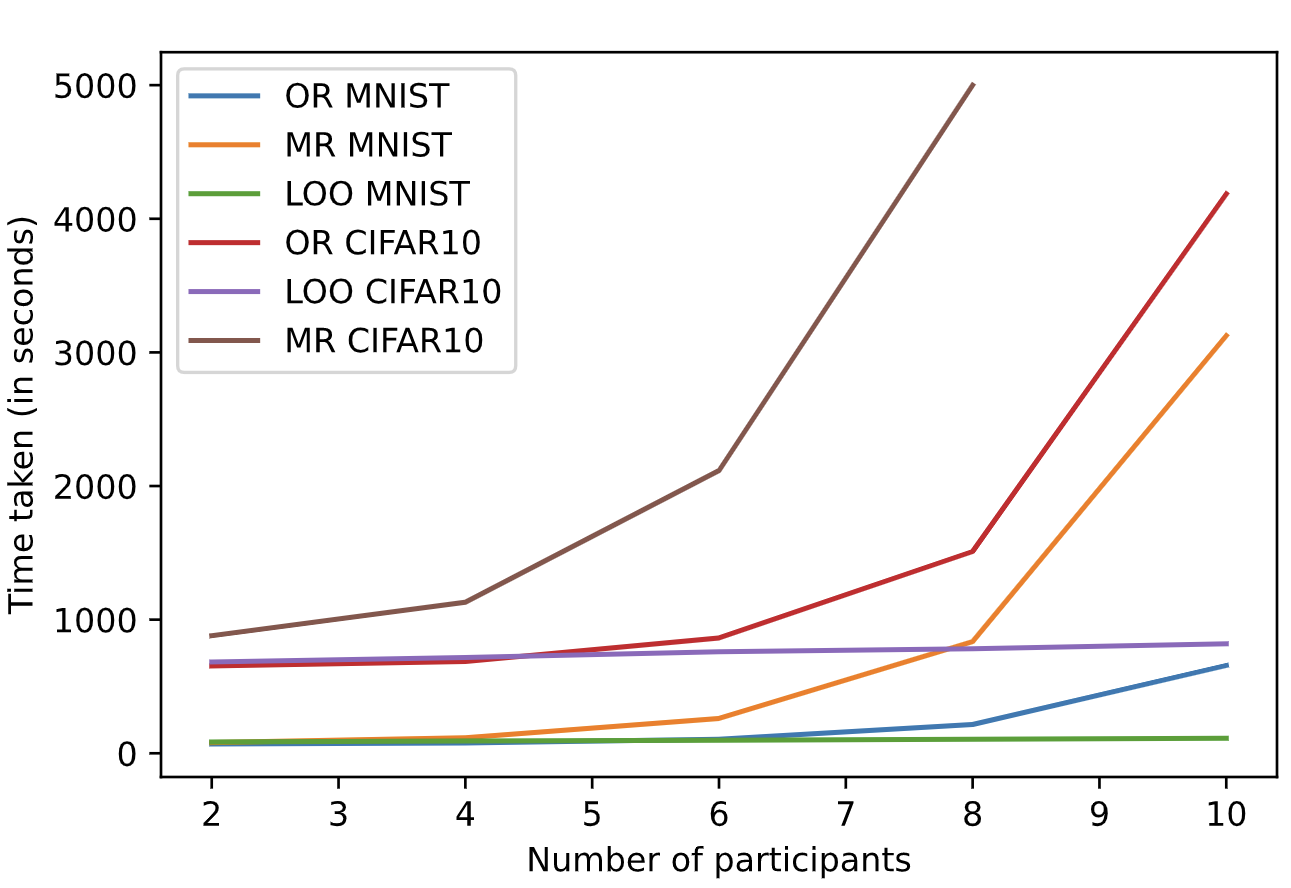}
    \caption{Comparison of Time Complexity.}
\end{figure}

\subsection{Additional metrics from experiments}
Apart from the CE values themselves, we present here the final global accuracy since the test accuracy is our utility function, the maximum difference between two elements in every payoff vector, the Euclidean distance between each payoff vector and an uninformative equal split of the value, and the total computation time. Since we are not concerned with a final model performance, we use the accuracy on the test set as the utility function and do not separate a validation set.

\begin{table}[]
\centering
\caption{Additional results on MNIST; Acc is the global model accuracy. Max Dif is the maximum distance between elements of the payoff vector. Distance refers to the Euclidean distance from the corresponding uniform vector. The time t is given in seconds. B is the total budget allocated by the un-normalized payoffs, and the only metric calculated before they are normalized.}
\label{tab:my-table}
\resizebox{\textwidth}{!}{%
\begin{tabular}{@{}lcllllllllllllllllllllllll@{}}
\toprule
{\ul \textbf{Participants}} & \multicolumn{5}{c}{2} & \multicolumn{5}{c}{4} & \multicolumn{5}{c}{6} & \multicolumn{5}{c}{8} & \multicolumn{5}{c}{10} \\ \midrule
\multicolumn{1}{|c}{} & \textbf{Acc} & \multicolumn{1}{c}{Max Dif} & \multicolumn{1}{c}{Dist} & \multicolumn{1}{c}{t} & \multicolumn{1}{c}{B} & \multicolumn{1}{c}{Acc} & \multicolumn{1}{c}{Max Dif} & \multicolumn{1}{c}{Dist} & \multicolumn{1}{c}{t} & \multicolumn{1}{c}{B} & \multicolumn{1}{c}{Acc} & \multicolumn{1}{c}{Max Dif} & \multicolumn{1}{c}{Dist} & \multicolumn{1}{c}{t} & \multicolumn{1}{c}{B} & \multicolumn{1}{c}{Acc} & \multicolumn{1}{c}{Max Dif} & \multicolumn{1}{c}{Dist} & \multicolumn{1}{c}{t} & \multicolumn{1}{c}{B} & \multicolumn{1}{c}{Acc} & \multicolumn{1}{c}{Max Dif} & \multicolumn{1}{c}{Dist} & \multicolumn{1}{c}{t} & \multicolumn{1}{c}{B} \\ \midrule
\textbf{OR - Shapley} & \multirow{2}{*}{94.9} & 0.92 & 0.65 & \multirow{2}{*}{69} & 0.9 & \multirow{2}{*}{93.31} & 0.93 & 0.7 & \multirow{2}{*}{76} & 0.8 & \multirow{2}{*}{92.29} & 0.68 & 0.55 & \multirow{2}{*}{105} & 0.8 & \multirow{2}{*}{91.6} & 0.58 & 0.54 & \multirow{2}{*}{215} & 0.5 & \multirow{2}{*}{90.93} & 0.5 & 0.49 & \multirow{2}{*}{657} & 0.8 \\ \cmidrule(r){1-1} \cmidrule(lr){3-4} \cmidrule(lr){6-6} \cmidrule(lr){8-9} \cmidrule(lr){11-11} \cmidrule(lr){13-14} \cmidrule(lr){16-16} \cmidrule(lr){18-19} \cmidrule(lr){21-21} \cmidrule(lr){23-24} \cmidrule(l){26-26} 
\textbf{OR - LC} &  & 0.74 & 0.52 &  & 1 &  & 0.53 & 0.5 &  & 0.9 &  & 0.37 & 0.41 &  & 0.9 &  & 0.29 & 0.36 &  & 0.9 &  & 0.24 & 0.33 &  & 0.9 \\ \midrule
\textbf{LOO no weights} & \multirow{3}{*}{94.95} & \textbf{0.98} & \textbf{0.69} & \multirow{3}{*}{84} & 2.4 & \multirow{3}{*}{93.3} & \textbf{0.98} & \textbf{0.84} & \multirow{3}{*}{91} & 1.3 & \multirow{3}{*}{92.29} & \textbf{1.04} & \textbf{0.9} & \multirow{3}{*}{97} & 0.4 & \multirow{3}{*}{91.6} & \textbf{0.82} & \textbf{0.73} & \multirow{3}{*}{105} & 0.2 & \multirow{3}{*}{90.9} & 0.56 & 0.52 & \multirow{3}{*}{112} & 0.1 \\ \cmidrule(r){1-1} \cmidrule(lr){3-4} \cmidrule(lr){6-6} \cmidrule(lr){8-9} \cmidrule(lr){11-11} \cmidrule(lr){13-14} \cmidrule(lr){16-16} \cmidrule(lr){18-19} \cmidrule(lr){21-21} \cmidrule(lr){23-24} \cmidrule(l){26-26} 
\textbf{LOO lin. weights} &  & \textbf{0.98} & \textbf{0.69} &  & 9.6 &  & \textbf{0.98} & \textbf{0.84} &  & 5.7 &  & 0.95 & 0.84 &  & 1.9 &  & 0.78 & 0.7 &  & 0.8 &  & \textbf{0.7} & \textbf{0.63} &  & 0.4 \\ \cmidrule(r){1-1} \cmidrule(lr){3-4} \cmidrule(lr){6-6} \cmidrule(lr){8-9} \cmidrule(lr){11-11} \cmidrule(lr){13-14} \cmidrule(lr){16-16} \cmidrule(lr){18-19} \cmidrule(lr){21-21} \cmidrule(lr){23-24} \cmidrule(l){26-26} 
\textbf{Reputation} &  & 0.22 & 0.15 &  & 1.6 &  & 0.18 & 0.18 &  & 2.9 &  & 0.2 & 0.2 &  & 3.2 &  & 0.15 & 0.16 &  & 5.4 &  & 0.14 & 0.18 &  & 6 \\ \midrule
\textbf{$\lambda$-MR} & \multirow{2}{*}{94.85} & 0.44 & 0.31 & \multirow{2}{*}{80} & 3.4 & \multirow{2}{*}{93.34} & 0.42 & 0.3 & \multirow{2}{*}{116} & 3.4 & \multirow{2}{*}{92.35} & 0.35 & 0.28 & \multirow{2}{*}{261} & 3.3 & \multirow{2}{*}{91.62} & 0.32 & 0.29 & \multirow{2}{*}{836} & 3.4 & \multirow{2}{*}{90.95} & 0.29 & 0.28 & \multirow{2}{*}{3125} & 3.3 \\ \cmidrule(r){1-1} \cmidrule(lr){3-4} \cmidrule(lr){6-6} \cmidrule(lr){8-9} \cmidrule(lr){11-11} \cmidrule(lr){13-14} \cmidrule(lr){16-16} \cmidrule(lr){18-19} \cmidrule(lr){21-21} \cmidrule(lr){23-24} \cmidrule(l){26-26} 
\textbf{Federated Shapley} &  & 0.56 & 0.39 &  & 4.2 &  & 0.52 & 0.38 &  & 4.1 &  & 0.41 & 0.33 &  & 4 &  & 0.37 & 0.33 &  & 4 &  & 0.32 & 0.31 &  & 4 \\ \bottomrule
\end{tabular}%
}
\end{table}
\begin{table}[]
\centering
\caption{Results on CIFAR-10; Acc is the global model accuracy. Max Dif is the maximum distance between elements of the payoff vector. Distance refers to the Euclidean distance from the corresponding uniform vector. The time t is given in seconds. B is the total budget allocated by the un-normalised payoffs, and the only metric calculated before they are normalized.}
\label{tab:my-table-cifar}
\resizebox{\textwidth}{!}{%
\begin{tabular}{@{}lcllllllllllllllllllllllll@{}}
\toprule
{\ul \textbf{Participants}} & \multicolumn{5}{c}{2} & \multicolumn{5}{c}{4} & \multicolumn{5}{c}{6} & \multicolumn{5}{c}{8} & \multicolumn{5}{c}{10} \\ \midrule
\multicolumn{1}{|c}{} & \textbf{Acc} & \multicolumn{1}{c}{Max Dif} & \multicolumn{1}{c}{Dist} & \multicolumn{1}{c}{t} & \multicolumn{1}{c}{B} & \multicolumn{1}{c}{Acc} & \multicolumn{1}{c}{Max Dif} & \multicolumn{1}{c}{Dist} & \multicolumn{1}{c}{t} & \multicolumn{1}{c}{B} & \multicolumn{1}{c}{Acc} & \multicolumn{1}{c}{Max Dif} & \multicolumn{1}{c}{Dist} & \multicolumn{1}{c}{t} & \multicolumn{1}{c}{B} & \multicolumn{1}{c}{Acc} & \multicolumn{1}{c}{Max Dif} & \multicolumn{1}{c}{Dist} & \multicolumn{1}{c}{t} & \multicolumn{1}{c}{B} & \multicolumn{1}{c}{Acc} & \multicolumn{1}{c}{Max Dif} & \multicolumn{1}{c}{Dist} & \multicolumn{1}{c}{t} & \multicolumn{1}{c|}{B} \\ \midrule
\textbf{OR - Shapley} & \multirow{2}{*}{0.84} & 0.62 & 0.44 & \multirow{2}{*}{653} & 0.71 & \multirow{2}{*}{0.78} & 0.46 & 0.34 & \multirow{2}{*}{687} & 0.68 & \multirow{2}{*}{0.73} & 0.42 & 0.34 & \multirow{2}{*}{863} & 0.63 & \multirow{2}{*}{0.68} & 0.46 & 0.42 & \multirow{2}{*}{1510} & 0.55 & \multirow{2}{*}{0.62} & 0.46 & 0.43 & \multirow{2}{*}{4186} & 0.53 \\ \cmidrule(r){1-1} \cmidrule(lr){3-4} \cmidrule(lr){6-6} \cmidrule(lr){8-9} \cmidrule(lr){11-11} \cmidrule(lr){13-14} \cmidrule(lr){16-16} \cmidrule(lr){18-19} \cmidrule(lr){21-21} \cmidrule(lr){23-24} \cmidrule(l){26-26} 
\textbf{OR - LC} &  & 0.31 & 0.22 &  & 0.84 &  & 0.41 & 0.33 &  & 0.78 &  & 0.3 & 0.24 &  & 0.73 &  & 0.23 & 0.22 &  & 0.68 &  & 0.17 & 0.21 &  & 0.62 \\ \midrule
\textbf{LOO no weights} & \multirow{3}{*}{0.837} & \textbf{0.94} & \textbf{0.66} & \multirow{3}{*}{683} & 1.4 & \multirow{3}{*}{0.78} & \textbf{0.55} & \textbf{0.41} & \multirow{3}{*}{717} & 1.4 & \multirow{3}{*}{0.74} & 0.38 & 0.34 & \multirow{3}{*}{760} & 1.1 & \multirow{3}{*}{0.68} & 0.33 & 0.33 & \multirow{3}{*}{782} & 1.05 & \multirow{3}{*}{0.65} & 0.35 & 0.4 & \multirow{3}{*}{819} & 0.68 \\ \cmidrule(r){1-1} \cmidrule(lr){3-4} \cmidrule(lr){6-6} \cmidrule(lr){8-9} \cmidrule(lr){11-11} \cmidrule(lr){13-14} \cmidrule(lr){16-16} \cmidrule(lr){18-19} \cmidrule(lr){21-21} \cmidrule(lr){23-24} \cmidrule(l){26-26} 
\textbf{LOO lin. weights} &  & 0.74 & 0.52 &  & 6.1 &  & 0.47 & 0.35 &  & 5.84 &  & 0.31 & 0.27 &  & 4.6 &  & 0.29 & 0.27 &  & 4.3 &  & \textbf{0.29} & \textbf{0.32} &  & 2.9 \\ \cmidrule(r){1-1} \cmidrule(lr){3-4} \cmidrule(lr){6-6} \cmidrule(lr){8-9} \cmidrule(lr){11-11} \cmidrule(lr){13-14} \cmidrule(lr){16-16} \cmidrule(lr){18-19} \cmidrule(lr){21-21} \cmidrule(lr){23-24} \cmidrule(l){26-26} 
\textbf{Reputation} &  & 0.25 & 0.17 &  & 1.6 &  & 0.11 & 0.1 &  & 3.4 &  & 0.125 & 0.14 &  & 4.8 &  & 0.17 & 0.16 &  & 5.8 &  & 0.14 & 0.16 &  & 7 \\ \midrule
\textbf{$\lambda$-MR} & \multicolumn{1}{l}{\multirow{2}{*}{0.832}} & 0.29 & 0.2 & \multirow{2}{*}{880} & 3.3 & \multirow{2}{*}{0.78} & 0.46 & 0.35 & \multirow{2}{*}{1130} & 3.4 & \multirow{2}{*}{0.74} & 0.48 & \textbf{0.4} & \multirow{2}{*}{2116} & 3.36 & \multirow{2}{*}{0.67} & \textbf{0.5} & \textbf{0.46} & \multirow{2}{*}{5000} & 3.36 & \multirow{2}{*}{-} & - & - & \multirow{2}{*}{-} & - \\ \cmidrule(r){1-1} \cmidrule(lr){3-4} \cmidrule(lr){6-6} \cmidrule(lr){8-9} \cmidrule(lr){11-11} \cmidrule(lr){13-14} \cmidrule(lr){16-16} \cmidrule(lr){18-19} \cmidrule(lr){21-21} \cmidrule(lr){23-24} \cmidrule(l){26-26} 
\textbf{Federated Shapley} & \multicolumn{1}{l}{} & 0.3 & 0.21 &  & 3.2 &  & 0.45 & 0.34 &  & 2.9 &  & \textbf{0.49} & \textbf{0.4} &  & 2.62 &  & \textbf{0.5} & \textbf{0.46} &  & 2.28 &  & - & - &  &  \\ \bottomrule
\end{tabular}%
}
\end{table}
\end{document}